\documentclass[11pt,letterpaper]{article}
\usepackage{emnlp2017}
\usepackage{times}
\usepackage{latexsym}

\usepackage{amsmath, amssymb}
\usepackage{graphicx}
\usepackage{color}
\usepackage{hyperref}
\usepackage{lscape}
\usepackage{verbatim}

\usepackage{url}
\newcommand{\our}{our method}
\newcommand{\mS}{{\mathcal S}}

\emnlpfinalcopy

\def\emnlppaperid{***}

\title{A Simple Approach to Learn Polysemous Word Embeddings}

\author{Yifan Sun \\
  Technicolor Research 
  \And
  Nikhil Rao\\
  Amazon\And
  Weicong Ding\\
  Amazon
  }

\date{}

\hypersetup{draft}
\begin{document}

\maketitle

\begin{abstract}
Many NLP applications require disambiguating polysemous words. 
Existing methods that learn polysemous word vector representations involve first detecting various senses
and optimizing the sense-specific embeddings separately,
which are invariably more involved than single sense learning methods such as word2vec. 
Evaluating these methods is also problematic, as rigorous quantitative evaluations in this space is limited, especially when compared with single-sense embeddings.
In this paper, we propose a simple method to learn a word representation, given {\it any} context. 
Our method only requires learning the usual single sense representation, and coefficients that can be learnt via a single pass over the data. 
We propose several new test sets for evaluating word sense induction, relevance detection, and  contextual word similarity, significantly supplementing the currently available tests.
Results on these and other tests show that while our method is embarrassingly simple, it achieves excellent results when compared to the state of the art models for unsupervised polysemous word representation learning. Our code and data are at \url{{https://github.com/dingwc/multisense/}} 
\end{abstract}

\section{Introduction}
\vspace*{-1ex}
Recent advances in word representation learning such as word2vec \cite{mikolov1} have significantly boosted the performance of numerous Natural Language Processing tasks \cite{mikolov1,  glove, levy2015improving}.
Despite their empirical performance, the inherent one-vector-per-word setting limits its application on tasks that require contextual understanding due to the existence of polysemous words such as part-of-speech tagging and semantic relatedness \cite{li2015multi}. 

To this end, various \emph{sense-specific word embeddings} have been proposed to account for the contextual subtlety of language \cite{ RMmultiple, RMmultiple2, ngmultiple, Mskipgram, TianProb2014,chen2014unified, li2015multi, linear_multisense}.
A majority of these methods propose to learn multiple vectors for each word via  clustering. \cite{RMmultiple, ngmultiple, Mskipgram} uses neural networks to learn cluster embeddings in order to matcha  polysemous word with its correct sense embeddings. 
Side information such as topical understanding \cite{liu2015topical, liu2015learning} or paralleled foreign language data \cite{guo2014learning, vsuster2016bilingual,iclr2017} have also been exploited for clustering different meanings of multi-sense words. 
Another trend is to forgo word embeddings in favor of sentence or paragraph embeddings for specific tasks
 \cite{amiri2016learning, kiros2015skip,le2014distributed}. 
 While being more flexible and adaptive to context, all these approaches require sophisticated neural network structures and are problem specific, taking away the advantage offered by the unsupervised embedding approaches of single-sense embeddings. This paper bridges this gap. 

In this paper we propose a novel and {\it extremely simple} approach to learn sense-specific word embeddings.
The essence of our approach is to assign each word a global base vector and model the contextual embedding as a {\it linear combination} of its context base vectors. 
Instead of a joint optimization to learn both base vector and combination weights, we propose to use the standard unisense word representation for the base vectors, and the (suitably normalized) word co-occurrence statistics as the linear combination weights; no further training computations are required in our approach. 

We evaluate our approach on various tasks that require \textit{contextual} understanding of words, combining existing and new test datasets and evaluation metrics: word-sense induction (\cite{koeling2005domain,bartunov2015breaking}), contextual word similarity (\cite{ngmultiple} and a new test set), and relevance detection (\cite{linear_multisense} and  a new test set). 
To the best of our knowledge, no prior literature has provided a comprehensive evaluation of all  these multisense-specific tasks. Our simple, intuitive model retains almost all  advantages offered by more complicated multisense embedding models, and often surpasses the performance of nonlinear  ``deep'' models. Our code and data are at \url{{https://github.com/dingwc/multisense/}}

To summarize, the contributions of our paper are as follows:
\begin{enumerate}
\item We propose an extremely simple model for learning polysemous word representations
\vspace*{-1ex}
\item We propose several new larger test sets to evaluate polysemous word embeddings, supplementing those that already exist
\vspace*{-1ex}
\item We perform extensive comparisons of our model to other widely used multisense models in the literature, and show that the simplicity of our model does {\textbf{not}} tradeoff performance
\end{enumerate}

The rest of the paper is organized as follows: in the next section, we introduce our model and provide a detailed explanation for obtaining the multisense word embeddings. In Section 3 qualitatively evaluate our model.
In Section 4 we introduce our new evaluation tasks and datasets in details. We also perform extensive experiments on four quantitative tests that are multisense specific. We finally conclude out paper in Section \ref{sec:conc}.

\section{Our Contextual Embedding Model}
\label{sec:model}
\vspace*{-1ex}
Like unisense vectors, sense-specific vectors should be closely aligned to words in that sense.
This idea of local similarity has been widely used to obtain context sense representation \cite{chen2014unified, ngmultiple, le2014distributed, Mskipgram}. It was also used to decompose unisense vector into sense specific vectors \cite{linear_multisense}.  
In this paper, we exploit this intuition and model the contextual embedding of a word as a linear combination of its contexts.

Specifically,  we consider a corpus drawn from a vocabulary $\mathcal{V} = (\text{word}_1,\hdots, \text{word}_V)$.
We define the \emph{normalized cooccurence matrix} as the $V\times V$ (sparse\footnote{The sparsity of $W$ was studied in \cite{glove}.}) symmetric matrix $W$ where 
\begin{equation}
\label{e-Wdef}
W_{ij} = \frac{\text{\# cooccurences of word$_i$, word$_j$}}{\text{freq. of word$_i$}\times \text{freq. of word$_j$}}.
\end{equation}
\vglue -1ex
{\noindent We }define a \emph{context} $\mathcal S$ as a collection of words provided alongside the target word. The context is flexible. It can be a sentence or a paragraph in which the word appeared, or a set of synonyms from WordNet. 
A standard unisense embedding (such as word2vec \cite{mikolov1}) can be represented as a  $d \times V$ matrix $C$, where $d$ is the {embedding dimension} and the $i$th column of $C$ is the embedding vector for the $i$th word in $\mathcal V$. 
Then the multisense embedding of word$_i$ given context $\mathcal S$ is
\begin{equation}
u =\frac{1}{|\mathcal S|} \sum_{j\mid \text{word}_j\in\mathcal S}W_{ji}C_j
\label{multisense}
\end{equation}
\vglue -1ex
{\noindent and} $C_j$ is the $j$th column of $C$. 
Take, for example, the  word \texttt{bank} with context {\it I must stop by the bank for a quick withdrawal}. The multisense embedding $u$ is a weighted sum of the base embeddings of each context word. Note that some words (\texttt{withdrawal}) are more relevant than others (\texttt{need, stop, quick}); the weight for each context word is the normalized co-occurance, which filters for relevant context words.

We can view the sets of columns of $C$ as spanning sense subspaces.
For example, the (likely low dimensional) subspace spanned by the submatrix of $C$ corresponding to vectors for  financial terms 
should also be highly correlated with 
 \texttt{savings}
 and much less correlated with \texttt{river};
 in other words, the mutisense word \texttt{bank} provided in either context will be well separated.

\paragraph{Implementation} In the remaining of the paper, we use $W$ via \eqref{e-Wdef},  constructed from  the 2016 English Wikipedia Dump\footnote{ {https://dumps.wikimedia.org/enwiki/}} with a local window of size $5$. The final vocabulary results after filtering away non-English words, stop words, and rare words occurring under 2,000 times. This results in a vocabulary of size $V=26974$. 
For base vectors $C$ we use either the pre-trained GLoVe embedding with $d=100$ \footnote{ {http://nlp.stanford.edu/projects/glove/}}, or the word2vec (w2v) embedding trained over the wikipedia corpus
with $d=50$ and $100$. The w2v embeddings are trained using skip-gram model with negative sampling. We set the number of negative samples to be $10$ and number of training epochs to be $15$.  
The $C$ and $W$ matrices are attached alongside our submission.


\section{Qualitative Examples}
\subsection{Norm Distribution of Our Approach}
\begin{figure}[hbt!]
\centering 
\includegraphics[width = 75mm]{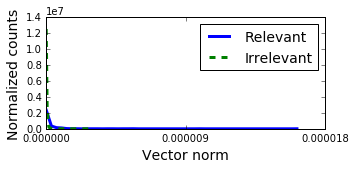}
\includegraphics[width = 70mm]{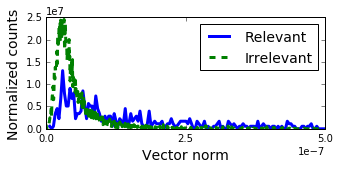} 
\caption{\small The norm distribution of contextual embeddings of all the word-context pairs in the word-context-relevance R1 dataset based on (uni=w2v,dim=100) embedding. The upper plot depicts the full distribution while the bottom plot depicts the zoomed-in distribution with $\ell_2$-norm close to $0$. Word-context pairs are labeled either relevant or irrelevant in the dataset. }
\label{f-norms-yawn}
\end{figure}
\begin{figure}[hbt!]
\centering 
\includegraphics[width = 70mm]{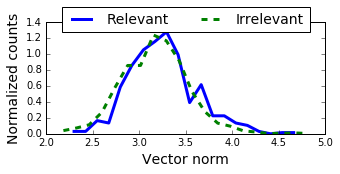} 
\includegraphics[width = 70mm]{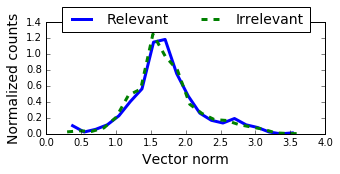} 
\includegraphics[width = 70mm]{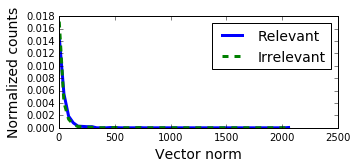} 
\caption{\small The norm distribution of all the word-context pairs in the word-context-relevance R1 dataset for \cite{Mskipgram} texttt{300d\_10s\_1.32c$\lambda$\_0mv} embedding  (top), \cite{ngmultiple} cossim embedding (middle), \cite{chen2014unified} (bottom). The norm-distribution plot indicates that contextual embeddings of relevant and irrelevant word-context pairs have similar norms, which is typical of existing polysemous word embeddings. Contrast this to Figure \ref{f-norms-yawn} (\our), where the norms are distributed according to relevance.}
\label{f-norms-others}
\end{figure}

  
Before we present qualitative examples, our first observation of the resulting embedding of word-context pairs is that the embedding vectors of words in irrelevant contexts have very low norms. In Fig.~\ref{f-norms-yawn} we selected around 500 word-context pairs from a word-context-relevance evaluation dataset (see Section 4 for details) and plot the histogram of the contextual embedding vector norms. In this evaluation data word-context pairs are labeled either relevant or irrelevant. 
As depicted in Fig.~\ref{f-norms-yawn}, the norm filtering effect is essential, given that unlike previous embeddings, our embeddings allow any word to act as context. In contrast, the multi-sense embeddings from prior literature (Figure \ref{f-norms-others}\footnote{Details of how we get embeddings from these competing methods are in Section 4.}) all have very similar norm distributions, regardless of context relevance.

This relevance filtering effect is advantageous in sentences where many neighboring words may not be describing at the query word.  However, the extreme decaying distribution of \our~(and \cite{chen2014unified}) in the above figures can make it difficult to measure contextual word similarity using simply cosine distance, as it magnifies words with very small norm that had already been identified as irrelevant. In the other extreme, using dot product overemphasizes common words. 
To mitigate this, we present a generalized similarity measure, with a tunable parameter $\alpha$ that describes exactly how much the norm should be taken into account. 

\subsection{Similarity Measure of Our Embedding} 
%

 
We propose to measure the contextual similarity using the geometric mean of the cosine similarity and dot product, as a tradeoff of these two extremes:
\begin{equation}
\label{eq:alpha}
d(x,y;\alpha) = \left(\frac{x^{\top}y}{\Vert x\Vert \Vert y \Vert}\right)^{\alpha} (x^{\top}y)^{1-\alpha}
\end{equation}
for $0\leq \alpha \leq 1$. Specifically, $d(x,y;1) = x^Ty$ and  $d(x,y;0)$ is the cosine distance between $x$ and $y$.

Table \ref{t-alpha-closest} looks at the closest words to \texttt{bank} in its two contexts for different choice of $\alpha$-distance measure. 
For dot-product we see an overemphasis of popular words that are only marginally related (\texttt{gently}, \texttt{steeply}).
For cosine similarity, rare words are overpromoted (\texttt{saxony}, \texttt{sacramento}).
In general, $\alpha = 0.75$ and $0.9$ can reasonably measure contextual similarity. 
\begin{table}[hbt!]
{\small
\begin{tabular}{|l|l|}
\hline
\multicolumn{2}{|l|}{ context:  \textit{institution	, currency, deposit,	money,	finance	}}\\\hline
$\alpha$&\emph{Closest words}\\\hline
 $0$ 
  & 
currencies, currency, deposit, laundering, franc
 \\\hline
$0.5$ & 
currencies, currency, deposit, laundering, franc
\\\hline
$0.75$ & 
deposit, currencies, currency, repayment, hedge
\\\hline
$0.9$ & 
repayment, cheque, deposit, liquidity, borrowers
\\\hline
 $1$  & 
bank, credit, triangle, saxony, linking
\\\hline
\multicolumn{2}{|l|}{ context:  \textit{water,	land,	sloping,	river,	flooding }}
\\\hline
$\alpha$ & Closest words\\\hline
 $0$  & 
steeply, gently, landslides, sloping, torrential
 \\\hline
$0.5$ & steeply, gently, landslides, tributaries, tributary\\\hline
$0.75$ &  tributaries, tributary, yangtze, confluence, empties\\\hline
$0.9$ & 
tributaries, tributary, yangtze, confluence, empties
\\\hline
 $1$  &  bank, sacramento, mouth, trail, fork
\\\hline
\end{tabular}}
\caption{ \small{Closest words to \texttt{bank} in the context of finance and geology, for various choices of $\alpha$ in \eqref{eq:alpha}. Recall that $\alpha = 0$ is the dot product and $\alpha = 1$ is cosine similarity.}}
\label{t-alpha-closest}
\end{table}

\begin{figure}[hbt!]
\label{f-alpha-sense}
\centering
\includegraphics[width = 75mm]{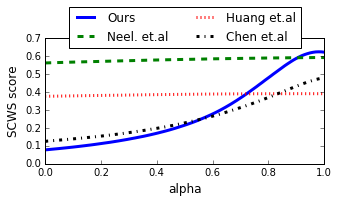}
\vspace*{-2ex}
\caption{\small{Spearman correlation score for varying $\alpha$ in \eqref{eq:alpha} on SCWS with different embeddings. Only the embedding that gave each approach the best SCWS scores are reported.}}
\end{figure}

To show the effect of $\alpha$ when only relevant $(w,\mS)$ pairs are used, Figure \ref{f-alpha-sense} plots the SCWS score (see Section \ref{sec:expts}; also \cite{ngmultiple}) for varying $\alpha$, for the top-performing embedding from each method. Cos-distance works best for all embeddings.
However, for embeddings from \cite{ngmultiple} and \cite{Mskipgram}, which all have norm $\approx$ 1, the choice of $\alpha$ makes little difference. 
On the other hand, using the embeddings from \cite{chen2014unified} and our method, which both have highly varying norms, the choice of $\alpha$ greatly affects performance.

\subsection{Qualitative Example}
Having explained the norm-filtering property of our approach and the $\alpha$-distance measure in Eq.~\eqref{eq:alpha}, we are now able to show a few qualitative examples of our model.
%
First,
Table~\ref{t-closestwords} shows closest words to \texttt{bank} in three different context. Here we used the GloVe embeddings as $C$ and set $\alpha = 0.9$. Contexts are selected  from news articles about finance, weather, and sports (an irrelevant context). The third case illustrates the filtering effect, with a norm that is an order of magnitude smaller than the first two.
\begin{table}[hbt!]
{\small
\begin{tabular}{|l|l|}
\hline
\multicolumn{2}{|l|}{\begin{minipage}[t]{0.45\textwidth}
 \it{Banco Santander of Spain said on Wednesday that its profit declined by   nearly half in the second quarter on restructuring charges and a contribution    to a fund to help finance bank bail ins in Europe.}
\end{minipage}}\\\hline
 norm &  $2.83\cdot 10^3$ \\
  neighbors &\begin{minipage}[t]{0.35\textwidth}
 banco, santander, hsbc, brasil, barclays
  \end{minipage} \\\hline\hline
  \multicolumn{2}{|l|}{
\begin{minipage}[t]{0.45\textwidth}
  \it{The Seine has continued to swell since the river burst its banks on  Wed., raising alarms throughout the city. As of 10pm on Friday its waters had reached 20 feet. The river was expected to crest on Sat. morning at up to  21 feet and to remain at high levels throughout the weekend.}
\end{minipage}}\\\hline
 norm&  $8.24\cdot 10^2$ \\ 
 neighborss &\begin{minipage}[t]{0.35\textwidth}
 banks, ganges, bank, tigris, river
\end{minipage}
 \\\hline\hline
\multicolumn{2}{|l|}{\begin{minipage}[t]{0.45\textwidth}
\it{Familia rebounded to strike out Tony Wolters, but first baseman James  Loney then fumbled a slow ground ball by pinch hitter Cristhian Adames  that allowed the tying run to score and left the bases loaded.}
\end{minipage}}\\\hline
 norm &  $8.78\cdot 10^1$ \\
 neighbor& \begin{minipage}[t]{0.35\textwidth}
 footed, balls, batsman, batsmen, winger
\end{minipage}
 \\\hline
\end{tabular}}
\vspace*{-2ex}
\caption{\small{Nearest words to and the $\ell_2$-norm of the contextual embedding of \texttt{bank} in 3 different contexts.}}
\label{t-closestwords}
\end{table}


\begin{table}[h]
{\small
\begin{tabular}{|l|l|}
\hline
\multicolumn{2}{|c|}{\it jack the ripper}\\\hline
 U &
 \begin{minipage}[t]{0.42\textwidth}
  billy, nicholson, tom, murphy, kelly 
 \end{minipage} \\
 M&
 \begin{minipage}[t]{0.42\textwidth}
 {\bf whitechapel}, {\bf murders},  judas, owens, whedon
  \end{minipage} \\\hline
\multicolumn{2}{|c|}{\it donald duck}\\\hline
 U&
 \begin{minipage}[t]{0.42\textwidth}
  jack, lee, george, lamb, howard 
   \end{minipage} \\
 M&
 \begin{minipage}[t]{0.42\textwidth}
\textbf{daffy},  lame, waterfowl, \textbf{goofy}, \textbf{scrooge}, \textbf{teal} 
   \end{minipage} 	\\\hline
   \multicolumn{2}{|c|}{\it steve jobs}\\\hline
 U & 
 \begin{minipage}[t]{0.42\textwidth}
  job, employees, hired, workers, manager 
   \end{minipage} \\
 M&
 \begin{minipage}[t]{0.42\textwidth}
 pixar,\textbf{ apple}, odd, \textbf{macintosh}, unemployed, commute 
   \end{minipage} 
\\\hline
\end{tabular}}
\vspace*{-2ex}
\caption{\small{Nearest words to  phrase embedding (excluding itself). Words relevant specifically to the phrase are bolded. U=unisense. M=multisense.}}
\label{t-phraseembedding}
\end{table}
Next, we investigate the use of our embedding model on phrase embeddings, constructed for example by averaging the embeddings of all words in the phrase, with the phrase itself as the context. 
In Table \ref{t-phraseembedding}, we pick three well-known bi-grams (\texttt{the} is a stopword and ignored). The bi-gram embedding is the average of either the unisense GloVe embeddings (U) or the multisense embedding (M) using our model. The closest words to these embeddings are listed in Table \ref{t-phraseembedding}.
We observe that the contextual phrase embeddings are able to pull out meanings having to do with the phrase as a whole, and not just the sum of its parts.


Finally, in Table \ref{t-filtering}, we list the words with highest norms, when projected in a single word context. In all the cases we observe,  high-norm words are highly relevant to the single context words. In the case of multisense words, a mixture of the different senses appear. (e.g., \texttt{chips} have \texttt{potato} and  \texttt{pentium} as relevant, keyboard has \texttt{layout} and \texttt{harpsichord}.)
\begin{table}[h]
{\small
\begin{tabular}{|l|l|}\hline
context & largest norm words \\\hline
eye & retina, ophthalmology, eye, sockets \\
keyboard & keyboard, layouts, harpsichord, sonata \\
run & yd, inning, td, rb \\
ball & fumbled, lucille, ball, wrecking \\
chips & chips, potato, pentium, chip \\
\hline
\end{tabular}}
\caption{\small{Top words in the context of a single word.}}
\label{t-filtering}
\end{table}

\section{Empirical study}
\label{sec:expts}

In this section we validate the performance of our embedding approach on a wide range of   tasks that {\it explicitly} require contextual understanding of words.
In Table \ref{t-quantitative} we collect (to our knowledge) the extent of multisense quantitative evaluations (columns 3-6 and 11-13), and supplement them with new, larger test sets  (columns 7-9). 
All tests are provided in the attached dataset folder.
In order to keep the evaluation fair across all embeddings, any word  that is not in the intersection of the vocabularies of all embeddings is removed from the tests; for the preexisting test sets,  this results in  slightly smaller test sets than those first proposed.

\paragraph{Similarity measure}
When irrelevant words are present, using $\alpha = 0.9$ is essential to leverage the norm distribution filtering effect. However, 
in all standard word similarity tasks, only relevant words are used as comparison. Therefore, in order to have our evaluations comparable with standard metrics, we keep $\alpha = 1$ (measuring cosine similarity for all tasks).

We compare our method against multisense approaches in \cite{ngmultiple, Mskipgram, chen2014unified}.
In each case, we use their pre-trained model and  choose the  embedding of a target word that is closest to the context representation (as they suggest). Since the code in \cite{ngmultiple} allows choosing various distance functions, we pick all and report the best scores.  For \cite{Mskipgram, chen2014unified} we use the cosine distance as recommended.

{\noindent\bf Overall Performance}  Table \ref{t-quantitative} shows that our method  consistently outperforms \cite{ngmultiple, Mskipgram}. We note that \cite{chen2014unified} is learned using additional supervision from the WordNet knowledge-base in clustering; therefore, it achieves comparably much higher scores in WSR and CWS tasks in which the evaluation is also based on WordNet. 
We now describe each task in detail.
\begin{table*}[hbt!]
\begin{center}
{\small
\begin{tabular}{|l|c|cc|cc|cc|cc|c|cc|}\hline
Embeddings &dim. &\multicolumn{6}{|c|}{\textbf{WCR}} & \multicolumn{2}{|c|}{\textbf{CWS}} & \textbf{SCWS} & \multicolumn{2}{|c|}{\textbf{WSC}} \\\hline
   &&\multicolumn{2}{|c|}{R1} &\multicolumn{2}{|c|}{R2}   & \multicolumn{2}{|c|}{R3}& &&&C1 & C2\\\hline
   &&Sp. & P@1&Sp. & P@1&Sp. & P@1  & AUC &AP & Sp.&Acc  &Acc  \\\hline
\textbf{\cite{ngmultiple}} &&&&&&&&&&&&\\
Euc. Dist. &50	& 0.08 & 0.13 & 0.24 &0.31 & 0.37 &0.45  &0.73&0.51 & 0.35& 0.72 & 0.60\\
Max Diff. 	 &50& 0.07 & 0.13 & 0.18 &0.25 & 0.29 & 0.38 &  0.73&0.52 & 0.32& 0.67 & 0.60\\
Min Diff. 	 &50& 0.01 & 0.09 & 0.02 &0.10 & 0.01 & 0.17  & 0.71&0.53 & 0.27 & 0.61& 0.60\\
Intersect dist.  &50& 0.02 & 0.36 & 0.10  &	0.46 & 0.07 &0.46 &  0.69&0.47 & 0.35 &0.62 & 0.60\\
Angle (cos-sim) &50 & 0.19 & 0.29 & 0.24 &0.33  & 0.34 &0.44 &  0.73&0.51 & 0.39 & 0.72 & 0.60\\
City block dist.  	 &50& 0.08 & 0.13 & 0.22 &	0.30 & 0.35 & 0.43 & 0.73&0.51& 0.36&0.68 & 0.60\\
Hamming dist. 	 &50& 0.15 & 0.27 & 0.19 &	0.31 & 0.27 & 0.43 & 0.72 &0.51  & 0.37 & 0.68 & 0.60\\
Chi Sq.	 &50& 0.10 &	0.17 & 0.14 &	0.20 & 0.52 & 0.19 &0.72& 0.52  & 0.32 & 0.67 & 0.60\\
 \hline
\textbf{\cite{Mskipgram}} &&&&&&&&&&&&\\
    3s 30kmv  &50& 0.20 & 0.27 & 0.25&	0.34  & 0.39 & 0.49  & 0.72 &0.47 & 0.53& 0.70 & 0.62\\
     3s 0mv   	 &300& 0.22 & 0.30 & 0.27&	0.38  & 0.41 & 0.54 &0.66 &	0.44  & 0.59 & 0.70 &0.62\\
     3s 30kmv    &300& 0.20 & 0.29 & 0.27&	0.39  & 0.42 & 0.53 & 0.69 & 0.45  & 0.58 & 0.70 &0.63\\
 10s 1.32c$\lambda$ 0mv &50& 0.21 & 0.29 & 0.25 & 0.35  & 0.43&0.55 &0.71 &0.47 & 0.53& 0.71 &0.63\\
 10s 1.32c$\lambda$ 30kmv &50& 0.20 & 0.30 & 0.24 & 0.30& 0.42& 0.52 & 0.72 &0.48 & 0.51& 0.69 &0.63\\
 10s 1.32c$\lambda$ 0mv &300& 0.22 & 0.32 & 0.27 &0.37 & 0.45 & 0.58 & 0.66& 0.44 &\bf{0.60} & 0.69& 0.63\\
\hline
\textbf{\cite{chen2014unified}} &200  & \bf{0.44} & \bf{0.73} & \bf{0.46} &	\bf{0.86} & \bf{0.63}& \bf{0.95} & \bf{0.96}& \bf{0.91} & 0.48& 0.75 & 0.66\\
\hline
 \bf{Our method}&&&&&&&&&&&&\\
   uni=GloVe  &100& \bf{0.34}	& \bf{0.54} & \bf{0.33}  & 0.51  & \bf{0.46} & \bf{0.61} & 0.89 & \bf{0.82} & 0.57 & \bf{0.83} & {\bf 0.77}\\
   uni=w2v &50& \bf{0.33} &	\bf{0.51}  & \bf{0.33} & \bf{0.52}  & \bf{0.46} & \bf{0.61} & \bf{0.89} &0.82& \bf{0.61} &  \bf{0.81} & {\bf 0.76}\\
   uni=w2v &100& 0.33 &0.51 & 0.33 & \bf{0.52} & 0.46 & 0.61  & \bf{0.90} & \bf{0.83}& \bf{0.62} &\bf{0.80} & {\bf 0.77}\\
\hline
\end{tabular}}
\vspace*{-2ex}
\caption{\small{Summary of all quantitative tests and performance metric of our embedding approach against compared baselines. For \cite{ngmultiple} different rows correspond to various types of distance used to get the contextual embedding. For \cite{Mskipgram} the row labels follow the terminology in the paper. We highlight the top 3 results for each test and metric in the table. Sp = Spearman correlation. P@1 = Precision@1. AP = Average Precision. AUC = Area Under Curve. Acc = Classification Accuracy. } 
}
\label{t-quantitative}
\end{center}
\end{table*}

{\noindent\bf Word-Context Relevance (WCR)} This task is proposed in \cite{linear_multisense} and aims to detect when word-context pairs are relevant.
In \cite{ngmultiple, Mskipgram,chen2014unified}, the relevance metric can be seen as the distance (cosine or Euclidean) between the query word and the context cluster center. 
In our method, we rely on the filtering effect of $W_{ij}$ values to diminish the norm of words in irrelevant contexts; thus  we propose the $\ell_2$-norms of the contextual embedding as the metric of relevance, where the target word is \emph{excluded} from the context if present.\footnote{The exclusion of the word is intended to highlight the filtering out of irrelevant contexts; in all other experiments (and as we intend in practice) the word is included in the context.}
In all cases, the ability of this metric to capture relevancy is essential for the success of that embedding to be applied to real world corpora, where not all neighboring words are relevant. 

The task is as follows. We have available to us some databases of words and their related words, which we view as that word's relevant context.
We create the ground-truth by setting the labels of related word-context pairs to be $1$, and for randomly picked word pairs  to be $0$. 
Specifically, R1 and R2 are constructed from the dataset in \cite{linear_multisense}, and R3 is a newly provided much larger test set, separately constructed from WordNet.

In R1, the negative samples are created by keeping the word unchanged and sampling $m=10$ random contexts. 
In R2, the context is unchanged, and $m=10$ random words are provided.
Note that for each word in each of the tests above, there is a single example with label $1$. 
In total, there are 137 words and 534 senses, with on average 6.98 context words per query word. Some examples are provided in Table \ref{t-WCRR1}.

R3 is a new test set that significantly augmnents R1 and R3. Here, we manually collect a set of $100$ polysemous words\footnote{https://en.wikipedia.org/wiki/List\_of\_true\_homonyms} and retrieve all their senses from  WordNet \cite{leacock1998combining}. We combine the definitions, synonyms, and examples sentences in WordNet as the context for each sense.  We have $566$ tests and for each test have $m=5$ negative samples, with random words in unchanged context. 
In total, there are 1938 words, 3234 senses, and on average 7.88 contexts per word, with some examples provided in Table {t-WCRR3}.

For each valid pair, we measure the {\it Spearman correlation} (Sp.) between relevance metrics and  ground-truth labels, as well as the {\it Precision@1}, {\it i.e.} the fraction of tests where the top item based on predicted scores is the valid pair. The reported performance metrics are averaged over all valid word-context pairs.

\begin{table}[htb!]
\centering
{\small
\begin{tabular}{|c|c|p{0.6\linewidth}|}
\hline 
Relevant? & Word & Context \\ 
\hline 
\multicolumn{3}{|l|}{Example test 1}\\ 
\hline 
True & tie & neck	men	front	worn	collar	knot	cloth	decorative \\ 
\hline 
False & tie & throw	propel	ball	direction	basket	goal	game	attain \\ 
\hline 
\multicolumn{3}{|l|}{Example test 2}\\ 
\hline 
True & tie & winner	score	tied	completion	identical	results	sports \\ 
\hline 
False & tie & domestic	hog	pig	culinary	eaten	cooked	fat \\ 
\hline 
\end{tabular}
\caption{Two example WCR-R1 tests for two senses of word \texttt{tie}. One valid (positive) word-context pair and a random (negative) pair are provided for each test.}
}
\label{t-WCRR1}
\end{table}
%


\begin{table}[htb!]
\centering
{\small
\begin{tabular}{|c|c|p{0.6\linewidth}|}
\hline 
Relevant? & Word & Context \\ 
\hline 
\multicolumn{3}{|l|}{Example test 1}\\  
\hline 
True & bank & slope turn road track higher inside order reduce effects force \\ 
\hline 
False & purposes & slope turn road track higher inside order reduce effects force \\ 
\hline 
\multicolumn{3}{|l|}{Example test 2}\\  
\hline 
True & bank &  
financial institution accepts deposits channels money activities check holds home \\
 \hline 
False & lip &  
financial institution accepts deposits channels money activities check holds home
 \\\hline 
\end{tabular}
\caption{Two example WCR-R3 tests for two valid senses of word \texttt{bank}. One valid (positive) word-context pair and a random (negative) pair are provided for each test.}
\label{t-WCRR3}
}
\end{table}

%
%

\begin{figure}[hbt!]
\centering 
\includegraphics[width = 70mm]{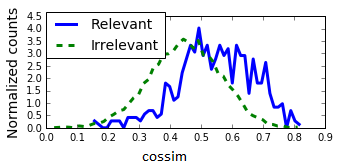}
\includegraphics[width = 70mm]{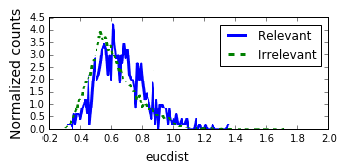} 
\includegraphics[width = 70mm]{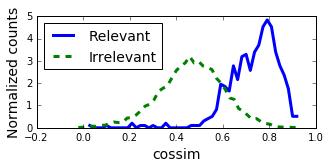} 
\caption{The distribution of cosine/Euclidean distances between (ir)relevant word and context cluster centers in the WCR R1 test for \cite{Mskipgram} texttt{300d\_10s\_1.32c$\lambda$\_0mv} embedding  (top), \cite{ngmultiple} cossim embedding (middle), and \cite{chen2014unified} (bottom). The amount of separation between the two distributions seems correlated with the success of the embedding on the WCR tasks.  In comparison, \our uses vector norms to distinguish relevance, the separation of which is shown in Figure \ref{f-norms-yawn}.}
\label{f-wcr-others}
\end{figure}

To further visualize the performance of each embedding on this task, we plot  the distribution of the relevance metric for relevant and Figure \ref{f-wcr-others} for the compared methods. It is clear that \cite{chen2014unified} has the best separation, corresponding to the highest score in the WCR columns of Table \ref{t-quantitative}. This can partially be explained by the fact that the construction of the embedding of \cite{chen2014unified} uses WordNet, which is also used in the construction of these tests.\footnote{R1 and R2 are constructed using a mixture of WordNet and human judgement; see \cite{linear_multisense}.}

{\noindent\bf Stanford's Contextual Word Similarities (SCWS)} 
%
The most popular existing test set for evaluating contextual word embedding similarity is the SCWS test \cite{ngmultiple}, which contains  $2003$ tests, each consisting of two word-context pairs $(w_1, S_1)$ and $(w_2, S_2)$. At all times, $S_1$ is relevant to $w_1$, and $S_2$ to $w_2$, but $w_1$ in the context of $S_1$ may not be synonymous with $w_2$ in the context of $S_2$. An example is given in Table \ref{t-scws-example}. In our evaluation, we first prune the test set to only include words present in vocabularies available to all embeddngs. Following \cite{ngmultiple},  we sort all the $n=2003$ test pairs based on predicted similarity score and compare such ranking against the ground-truth ranking indicated by the average human evaluation score. The distance between two rank-lists is measured using the Spearman correlation score. 

\begin{table}[htb!]
\centering
{\small
\begin{tabular}{|p{\linewidth}|}
\hline 
Example of SCWS test for \texttt{admission} and \texttt{confession} \\ 
\hline 
   ... the reason the Buddha himself gave was that the \emph{admission} of women would weaken the Sangha and shorten its lifetime ... \\ 
\hline 
 ... They included a \emph{confession} said to have been inadvertently included on a computer disk that was given to the press...
\\ 
\hline 
avg. human-given sim. score:   2.3\\
\hline
\end{tabular}
\caption{An example of a pair of word-contexts for a single SCWS task.}
}
\label{t-scws-example}
\end{table}

We note that in \cite{ngmultiple} the similarity between two word-context pairs is the measured using {\it avgSimC}, a weighted average of cosine similarities between all possible representation vectors of $w_1$ and $w_2$. This metric, however, can not be applied to our approach since we have an infinite number of possible contextual representation for each word. Therefore, we use the cosine similarity without averaging, which is reasonable for all the embedding approaches. We note that the cosine similarity is also used in \cite{Mskipgram, RMmultiple2}
Of course, this is disadvantageous for the embeddings of \cite{ngmultiple}, and our scores of their embedding are closer to that reported in \cite{Mskipgram}, which also does not use averaging.


{\noindent\bf Our Contextual Word Similarity (CWS)} 
We expand upon the SCWS test by providing our own larger CWS test, constructed mostly in an unsupervised manner based on WordNet. 
We retrieve a set of multisense words and their senses from WordNet as in {\bf WCR} R3, with contexts as the concatenation of the definition and all example sentences. 
The full list of tests are attached in the dataset folder, with an example  in Table~\ref{t-CWS}. 
Note that compared to the SCWS test set, the contexts are much shorter and less noisy. 

For each  query word-context $(w,\mS)$ pair, we attach a positive label to another $(w',\mS')$ if $\mS'=\mS$ and $w$, $w'$ are similar words. 
We collect negative samples as pairs $(w'',\mS'')$ if $\mS''\neq \mS$, where either $w = w''$ or $w$ and $w''$ are marked similar words by WordNet in the context $\mS''$. 
Given a query $(w,\mS)$ pair, the goal is to rank the  similar (positive) pairs above the dissimilar (negative) ones in the context of $\mS$. In all, we create a set of $3,955$ tests based on $154$ polysemous words. We calculated the cosine similarity between the contextual embeddings of negative/positive samples and the query, and report scores in Table \ref{t-quantitative}.

\begin{table}
\centering
{\small
\begin{tabular}{|c|c|p{0.5\linewidth}|}
\hline 
Label & Word & Context \\ 
\hline 
query & coach & sports charge training athlete team
 \\ 
\hline 
True & manager & sports charge training athlete team \\ 
\hline 
False & bus & vehicle carrying passengers public transport rode  work\\ 
\hline 
False & coach & carriage pulled horses driver\\ 
\hline
\end{tabular}
\caption{Example test in Contextual Word Similarity dataset. }
}
\label{t-CWS}
\end{table} 

 {\noindent\bf Word-Sense Classification (WSC)} 
Both WCR and CWS tasks are heavily based on WordNet, and offer an unfair advantage to multisense embeddings whose construction is also based on WordNet.
In this sense, SCWS offers a more generalizable evaluation. Two additional  word-sense tests are 
and \cite{koeling2005domain,bartunov2015breaking}. 
Similar to the Word-Sense Induction (WSI) task provided by the same works, we devise a Word Sense Classification  (WSC) task, to predict the correct sense of a polysemous word in a given sentence or paragraph.

We construct the tests from sense-labeled $(w,S)$ pairs in \cite{koeling2005domain} (C1) and \cite{wsd_dataset}
 (C2) by merge all the training and test data 
and further remove rare senses with  $<10$ examples sentences.  Some examples of the C1 dataset are provided in Table~\ref{t-example-in-classification} (C2 is very similar).  
In total, C1 contains 39 words, 116 senses, and 11,064 lines. C2 is much bigger, containing 783 words, 5,188 senses, and 961,670 examples. Given that such large datasets were already available, we did not need to create our own.
 For each word we create $80-20\%$ train-test splits, train a $K$-NN multiclass classifier with Euclidean distance between contextual word embeddings,
 and report the {\it  mean classification accuracy} (Acc) averaged across all words.

\begin{table}[htb!]
\centering
{\small
\begin{tabular}{|c|c|p{0.425\linewidth}|}
\hline 
Word & Sense & Context \\ 
\hline 
\multicolumn{3}{|l|}{Example test 1}\\ 
\hline 
coach$\%$1:06:02:: & coach & It was perfect for low fare express coach services. \\ 
\hline 
coach$\%$1:18:01:: & coach & But Chicago coach Phil Jackson said the bulls had done a good team job of preventing Malone from getting the easy transition baskets he thrives on nn even when the lumbering longley could not keep up with him. \\ 
\hline 
coach$\%$1:06:02:: & coach & Police said special trains and coaches had already been booked in Belgium alone for Sunday's march for jobs. \\ 
\hline 
\multicolumn{3}{|l|}{Example test 2}\\
\hline 
right$\%$1:07:00::& right & The buyer can exercise this right by refusing to take delivery or informing the seller that he rejects the goods.
  \\\hline 
right$\%$10:15:00:: & right & Woods hit first and stuck his approach about feet to the right of the pin.
 \\\hline 
\end{tabular}
\caption{Example data (word, context sentence, and word-sense label) in Word-Sense Classification WSC-C1 dataset. The word-sense labels follow the format in WordNet.}
}
\label{t-example-in-classification}
\end{table}

 {\noindent\bf Discussion} 
We have provided a wide array of evaluations for measuring different aspects of multisense word  embeddings, both collected from existing test sources and formed ourselves through WordNet. Overall, we find that our simple model performs surprisingly well on all evaluations, with the only consistent competitor a WordNet based model. 

One thing to note is that the SCWS Spearman scores of the \cite{ngmultiple} listed here are much smaller than that first reported. This is entirely attributed to the fact that we use direct cosine similarity between word embeddings, whereas they use an averaged similarity across their provided context words. Both are perfectly valid metrics; our choice is solely so that the identical metric can be applied to \emph{all} embeddings, where this averaged similarity metric cannot be used.

%


\section{Conclusion}
\label{sec:conc}
In this paper, we developed a method that can yield contextual word embeddings for \em{any} word under \em{any} context. When the context is irrelevant to the word, the embedding norms will be almost 0. A key highlight of our method is the simplicity, both from the modeling and the learning point of view. Experiments on several datasets and on several tasks show that the method we propose is competitive with the state of the art when it comes to unsupervised methods to learn polysemous word representations. 

\newpage
\bibliography{emnlp2017}

\begin{thebibliography}{}
\expandafter\ifx\csname natexlab\endcsname\relax\def\natexlab#1{#1}\fi

\bibitem[{Arora et~al.(2016)Arora, Li, Liang, Ma, and
  Risteski}]{linear_multisense}
Sanjeev Arora, Yuanzhi Li, Yingyu Liang, Tengyu Ma, and Andrej Risteski. 2016.
\newblock Linear algebraic structure of word senses, with applications to
  polysemy.
\newblock {\em arXiv preprint arXiv:1601.03764\/} .

\bibitem[{Bartunov et~al.(2015)Bartunov, Kondrashkin, Osokin, and
  Vetrov}]{bartunov2015breaking}
Sergey Bartunov, Dmitry Kondrashkin, Anton Osokin, and Dmitry Vetrov. 2015.
\newblock Breaking sticks and ambiguities with adaptive skip-gram.
\newblock {\em arXiv preprint arXiv:1502.07257\/} pages 47--54.

\bibitem[{Chen et~al.(2014)Chen, Liu, and Sun}]{chen2014unified}
Xinxiong Chen, Zhiyuan Liu, and Maosong Sun. 2014.
\newblock A unified model for word sense representation and disambiguation.
\newblock In {\em EMNLP\/}. Citeseer, pages 1025--1035.

\bibitem[{Dhillon et~al.(2015)Dhillon, Foster, and
  Ungar}]{dhillon2015eigenwords}
Paramveer Dhillon, Dean Foster, and Lyle Ungar. 2015.
\newblock Eigenwords: Spectral word embeddings.
\newblock {\em Journal of Machine Learning Research\/} pages 3035 -- 3078.

\bibitem[{Guo et~al.(2014)Guo, Che, Wang, and Liu}]{guo2014learning}
Jiang Guo, Wanxiang Che, Haifeng Wang, and Ting Liu. 2014.
\newblock Learning sense-specific word embeddings by exploiting bilingual
  resources.
\newblock In {\em COLING\/}. pages 497--507.

\bibitem[{Huang et~al.(2012)Huang, Socher, Manning, and Ng}]{ngmultiple}
Eric~H Huang, Richard Socher, Christopher~D Manning, and Andrew~Y Ng. 2012.
\newblock Improving word representations via global context and multiple word
  prototypes.
\newblock In {\em Proceedings of the 50th Annual Meeting of the Association for
  Computational Linguistics: Long Papers-Volume 1\/}. Association for
  Computational Linguistics, pages 873--882.

\bibitem[{Kiros et~al.(2015)Kiros, Zhu, Salakhutdinov, Zemel, Urtasun,
  Torralba, and Fidler}]{kiros2015skip}
Ryan Kiros, Yukun Zhu, Ruslan~R Salakhutdinov, Richard Zemel, Raquel Urtasun,
  Antonio Torralba, and Sanja Fidler. 2015.
\newblock Skip-thought vectors.
\newblock In {\em Advances in neural information processing systems\/}. pages
  3294--3302.

\bibitem[{Koeling et~al.(2005)Koeling, McCarthy, and
  Carroll}]{koeling2005domain}
Rob Koeling, Diana McCarthy, and John Carroll. 2005.
\newblock Domain-specific sense distributions and predominant sense
  acquisition.
\newblock In {\em Proceedings of the conference on Human Language Technology
  and Empirical Methods in Natural Language Processing\/}. Association for
  Computational Linguistics, pages 419--426.

\bibitem[{Le and Mikolov(2014)}]{le2014distributed}
Quoc~V Le and Tomas Mikolov. 2014.
\newblock Distributed representations of sentences and documents.
\newblock In {\em ICML\/}. volume~14, pages 1188--1196.

\bibitem[{Leacock and Chodorow(1998)}]{leacock1998combining}
Claudia Leacock and Martin Chodorow. 1998.
\newblock Combining local context and wordnet similarity for word sense
  identification.
\newblock {\em WordNet: An electronic lexical database\/} 49(2):265--283.

\bibitem[{Levy and Goldberg(2014)}]{wvmf}
Omer Levy and Yoav Goldberg. 2014.
\newblock Neural word embedding as implicit matrix factorization.
\newblock In {\em Advances in Neural Information Processing Systems\/}. pages
  2177--2185.

\bibitem[{Levy et~al.(2015)Levy, Goldberg, and Dagan}]{levy2015improving}
Omer Levy, Yoav Goldberg, and Ido Dagan. 2015.
\newblock Improving distributional similarity with lessons learned from word
  embeddings.
\newblock {\em Transactions of the Association for Computational Linguistics\/}
  3:211--225.

\bibitem[{Li and Jurafsky(2015)}]{li2015multi}
Jiwei Li and Dan Jurafsky. 2015.
\newblock Do multi-sense embeddings improve natural language understanding?
\newblock {\em arXiv preprint arXiv:1506.01070\/} .

\bibitem[{Liu et~al.(2015{\natexlab{a}})Liu, Qiu, and Huang}]{liu2015learning}
Pengfei Liu, Xipeng Qiu, and Xuanjing Huang. 2015{\natexlab{a}}.
\newblock Learning context-sensitive word embeddings with neural tensor
  skip-gram model.
\newblock In {\em Twenty-Fourth International Joint Conference on Artificial
  Intelligence\/}.

\bibitem[{Liu et~al.(2015{\natexlab{b}})Liu, Liu, Chua, and
  Sun}]{liu2015topical}
Yang Liu, Zhiyuan Liu, Tat-Seng Chua, and Maosong Sun. 2015{\natexlab{b}}.
\newblock Topical word embeddings.
\newblock In {\em AAAI\/}. pages 2418--2424.

\bibitem[{Mikolov et~al.(2013{\natexlab{a}})Mikolov, Chen, Corrado, and
  Dean}]{mikolov2}
Tomas Mikolov, Kai Chen, Greg Corrado, and Jeffrey Dean. 2013{\natexlab{a}}.
\newblock Efficient estimation of word representations in vector space.
\newblock {\em arXiv preprint arXiv:1301.3781\/} .

\bibitem[{Mikolov et~al.(2013{\natexlab{b}})Mikolov, Sutskever, Chen, Corrado,
  and Dean}]{mikolov1}
Tomas Mikolov, Ilya Sutskever, Kai Chen, Greg~S Corrado, and Jeff Dean.
  2013{\natexlab{b}}.
\newblock Distributed representations of words and phrases and their
  compositionality.
\newblock In {\em Advances in neural information processing systems\/}. pages
  3111--3119.

\bibitem[{Neelakantan et~al.(2015)Neelakantan, Shankar, Passos, and
  McCallum}]{Mskipgram}
Arvind Neelakantan, Jeevan Shankar, Alexandre Passos, and Andrew McCallum.
  2015.
\newblock Efficient non-parametric estimation of multiple embeddings per word
  in vector space.
\newblock {\em arXiv preprint arXiv:1504.06654\/} .

\bibitem[{Pennington et~al.(2014)Pennington, Socher, and Manning}]{glove}
Jeffrey Pennington, Richard Socher, and Christopher~D Manning. 2014.
\newblock Glove: Global vectors for word representation.
\newblock In {\em EMNLP\/}. volume~14, pages 1532--1543.

\bibitem[{Raganato et~al.(2017)Raganato, Camacho-Collados, and
  Navigli}]{wsd_dataset}
Alessandro Raganato, Jose Camacho-Collados, and Roberto Navigli. 2017.
\newblock Word sense disambiguation: A unified evaluation framework and
  empirical comparison.
\newblock In {\em Proceedings of EACL\/}. Valencia, Spain.

\bibitem[{Reisinger and Mooney(2010{\natexlab{a}})}]{RMmultiple2}
Joseph Reisinger and Raymond Mooney. 2010{\natexlab{a}}.
\newblock A mixture model with sharing for lexical semantics.
\newblock In {\em Proceedings of the 2010 Conference on Empirical Methods in
  Natural Language Processing\/}. EMNLP '10, pages 1173--1182.

\bibitem[{Reisinger and Mooney(2010{\natexlab{b}})}]{RMmultiple}
Joseph Reisinger and Raymond~J. Mooney. 2010{\natexlab{b}}.
\newblock Multi-prototype vector-space models of word meaning.
\newblock In {\em The 2010 Annual Conference of the North American Chapter of
  the Association for Computational Linguistics\/}. pages 109--117.

\bibitem[{Shyam et~al.(2017)Shyam, Kai-Wei, James, Matt, and Adam}]{iclr2017}
Upadhyay Shyam, Chang Kai-Wei, Zou James, Taddy Matt, and Kala Adam. 2017.
\newblock Beyond bilingual: Multi-sense word embeddings using multilingual
  context.
\newblock In {\em ICLR\/}.

\bibitem[{{\v{S}}uster et~al.(2016){\v{S}}uster, Titov, and van
  Noord}]{vsuster2016bilingual}
Simon {\v{S}}uster, Ivan Titov, and Gertjan van Noord. 2016.
\newblock Bilingual learning of multi-sense embeddings with discrete
  autoencoders.
\newblock {\em arXiv preprint arXiv:1603.09128\/} .

\bibitem[{Tian et~al.(2014)Tian, Dai, Bian, Gao, Zhang, Chen, and
  Liu}]{TianProb2014}
Fei Tian, Hanjun Dai, Jiang Bian, Bin Gao, Rui Zhang, Enhong Chen, and Tie-Yan
  Liu. 2014.
\newblock A probabilistic model for learning multi-prototype word embeddings.

\end{thebibliography}
\bibliographystyle{emnlp_natbib}

\end{document}